\documentclass[conference]{IEEEtran}
\IEEEoverridecommandlockouts

\usepackage{cite,comment}
\usepackage{amsmath,amssymb,amsfonts}
\usepackage{algorithm} 
\usepackage{algorithmic}
\usepackage{graphicx,booktabs}
\usepackage{textcomp}
\usepackage{xcolor}
\def\BibTeX{{\rm B\kern-.05em{\sc i\kern-.025em b}\kern-.08em
    T\kern-.1667em\lower.7ex\hbox{E}\kern-.125emX}}

\usepackage{tikz}
\usetikzlibrary{shapes.callouts, arrows.meta, positioning, backgrounds}
\usepackage{multirow}\usepackage{colortbl}
\usepackage{makecell}

\begin{document}

\title{LLM-Assisted Logic Rule Learning: Scaling Human Expertise for Time Series Anomaly Detection
}

\author{\IEEEauthorblockN{Haoting Zhang, Shekhar Jain}
\IEEEauthorblockA{\textit{Automated Inventory Management, Supply Chain Optimization Technologies} \\
\textit{Amazon}\\
Bellevue, USA \\
\{htizhang, shekhajt\}@amazon.com}

}

\maketitle

\begin{abstract}
Time series anomaly detection is critical for supply chain management to take proactive operations, but faces challenges: classical unsupervised anomaly detection based on exploiting data patterns often yields results misaligned with business requirements and domain knowledge, while manual expert analysis cannot scale to millions of products in the supply chain. We propose a framework that leverages large language models (LLMs) to systematically encode human expertise into interpretable, logic-based rules for detecting anomaly patterns in supply chain time series data. Our approach operates in three stages: 1) LLM-based labeling of training data instructed by domain knowledge, 2) automated generation and iterative improvements of symbolic rules through LLM-driven optimization, and 3) rule augmentation with business-relevant anomaly categories supported by LLMs to enhance interpretability. The experiment results showcase that our approach outperforms the unsupervised learning methods in both detection accuracy and interpretability. Furthermore, compared to direct LLM deployment for time series anomaly detection, our approach provides consistent, deterministic results with low computational latency and cost, making it ideal for production deployment. The proposed framework thus demonstrates how LLMs can bridge the gap between scalable automation and expert-driven decision-making in operational settings.

\end{abstract}


\section{Introduction}
\textbf{Background.} Time series anomaly detection is critical for monitoring large-scale operational systems, from cloud infrastructure health to manufacturing quality control to supply chain management \cite{sun2024unraveling,zhang2024daily,obata2025robust}. Early detection of irregularities enables proactive intervention before issues escalate into costly disruptions or service degradation. This paper focuses on Amazon's supply chain management, where continuous monitoring of weekly metrics at the individual product level (ASINs\footnote{
An Amazon ASIN (Amazon Standard Identification Number) is a unique, 10-character alphanumeric code assigned by Amazon to every product in its catalog, serving as an identifier for product management, inventory, and customer search.}) is essential for maintaining optimal inventory, detecting demand anomalies, and enabling timely corrective actions. Given Amazon's scale with hundreds of millions of individual products across diverse categories and marketplaces, an automated and efficient anomaly detection system is indispensable for transforming reactive problem-solving into proactive supply chain management \cite{liu2025easyad,yao2025moon,liu2025tsb}.

\textbf{Baselines \& Challenges.}
Time series anomaly detection has largely been approached as an unsupervised learning problem due to the scarcity and cost of labeled data \cite{munir2018deepant,audibert2020usad,zhao2022comparative}. Thus, statistical and learning-based methods have been widely employed to exploit data patterns to detect anomalies by identifying deviations from established norms or trends. A partial list of methods includes statistical methods (z-scores, ARIMA), tree-based isolation mechanisms (Isolation Forest \cite{4781136}), and machine learning algorithms/models (Long-Short-Term Memory networks, Variational Autoencoders, Transformers, etc.) \cite{tedjopurnomo2020survey,yang2023dcdetector,chen2023imdiffusion,jhin2023precursor,zamanzadeh2024deep,zhan2024meta,fang2024temporal,zhang2024does,zhou2025crosslinear,zhou2025mtsbench}. For example, Amazon's in-production method for category-level anomaly detection, rather than individual products, employs an Isolation Forest-based method and has proven to be effective at aggregated category levels, where aggregated time series data exhibit tractable trending patterns for the anomaly detection method to learn.


On the other hand, these unsupervised pattern-based anomaly detection methods face fundamental challenges at the level of individual products. First, individual product time series exhibit extreme volatility and uncertainty, with data oscillating irregularly and providing insufficient stable patterns for these methods to learn reliable detection rules. Consequently, Amazon's in-production method for anomaly detection at aggregated category levels can hardly distinguish genuine anomalies from natural fluctuations in high-volatility individual product data. Second, these methods are generally blind to application-specific contexts or business logic: a sharp decrease in an out-of-stock ratio indicates an improvement and should not be flagged, yet pattern-based methods detect it as an anomaly because it deviates from historical norms or trends. Furthermore, even if these methods achieved acceptable accuracy, their black-box nature provides no interpretable rationale for detected anomalies, hindering operational decision-making and validation by domain experts \cite{srivastava2020forecasting,schneider2021distributed,ahmed2021anomaly,paparrizos2022tsb,paparrizos2022volume,han2022deeptea,liang2025kdselector,gu2025argos}.

In contrast, human domain experts effectively identify anomalies in individual product time series data by integrating visual pattern recognition with contextual business knowledge. Human judgment incorporates both data patterns (recognizing irregular spikes or trends) and application contexts (understanding when deviations are problematic versus beneficial). However, manual expert analysis is prohibitively expensive and time-consuming, making it impractical for monitoring across hundreds of millions of individual products. Recent work has explored using large language models (LLMs) to replicate human-like reasoning for time series analysis and delivered promising results \cite{gruver2023large,jin2023time,chandrayanlead,zhou2024can,dong2024can,tsai2025anollm}. However, direct LLM-based detection for millions of time series suffers from high computational latency and costs, non-deterministic outputs that lack reproducibility, and potential hallucinations that undermine reliability in production systems. This motivates our central research question: \emph{How can we scale human 
expertise to large-scale time series anomaly detection while avoiding direct LLM deployment in production?}

\begin{figure}
    \centering
    \includegraphics[width=\linewidth]{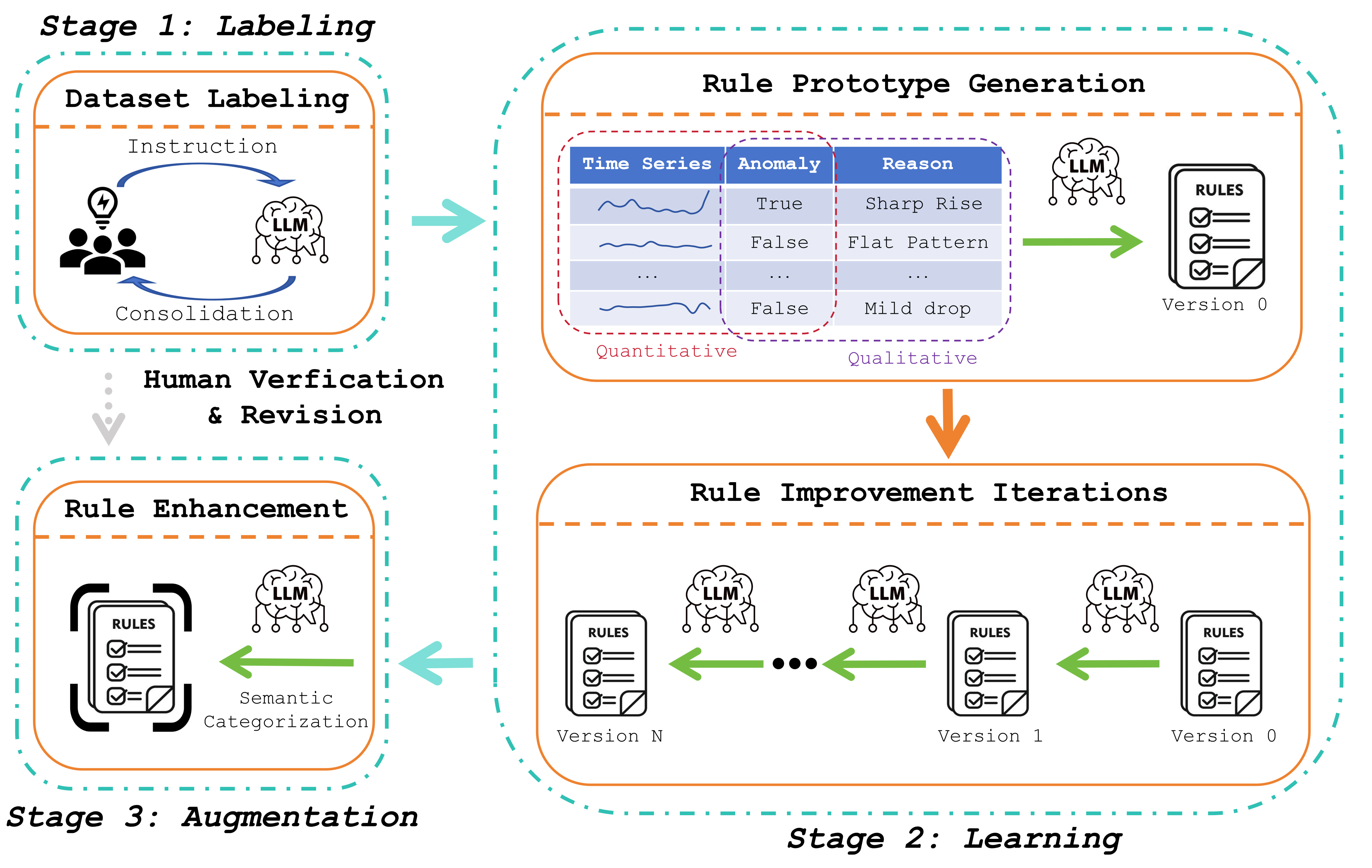}
    \caption{A summarized three-stage framework of our approach: 1) instructing multimodal (vision-language) LLMs to label a training dataset, 2) LLM-driven automated learning of logic-based symbolic rules from the labeled dataset, and 3) employing LLMs to categorize detection rules to augment business interpretability. Our approach leverages LLMs to first encode human domain expertise into a labeled dataset and then distill it into logic-based rules.
}
    \label{fig:framework}
\end{figure}

\textbf{Our Approach.} Instead of direct use for anomaly detection in production, LLMs in our approach serve as a bridge that transforms implicit human domain expertise into explicit and interpretable logic-based symbolic rules that can be executed efficiently at large scales. Specifically, we propose LLM-assisted logic-based rule learning for detecting anomalies, which operates in three stages (Fig. \ref{fig:framework}). In the first ``labeling'' stage, we employ multimodal (vision-language) LLMs to automatically label a sufficiently large dataset by providing them with both the visualized time series data and the business context. The detection results delivered by LLMs contain not only the binary anomaly classification, but also the reasons for the decision. In this manner, business insights and expertise are encoded into the labeled dataset through the help of multi-modal LLMs. With the labeled dataset, we begin the second ``learning'' stage: learning logic-based rules for anomaly detection from the labeled dataset. In this stage, initial rules are generated by the LLM using both qualitative reasoning and quantitative thresholds extracted from the labeled dataset. Then, we use the LLM to iteratively improve these rules, with rule behavior analysis and associated targeted modifications. By regarding the targeted modification of the rule supported by the LLM as ``semantic gradient,'' we cast our iterative improvement pipeline into a standardized machine learning (ML) framework. This enables us to implement ML strategies such as early stopping and multi-start learning to avoid overfitting of the learned rules. After the rules are learned, we implement an ``augmentation'' stage that leverages LLM to categorize the detection rules into explainable categories, that further augments the interpretability of the rules for business applications. The learned rules possess explicit detection conditions for humans to verify and optionally revise in production.

\section{Methodology}
\label{sec.method}
The proposed large language model (LLM)-assisted rule learning for time series anomaly detection operates in three stages: 1) a labeling stage (Sec. \ref{sec:label}), 2) a learning stage (Sec. \ref{sec:learn}), and 3) an augmentation stage (Sec. \ref{sec:revision}).

\subsection{Labeling Stage}
\label{sec:label}
To incorporate human domain knowledge into a scalable and interpretable anomaly detection, we leverage the power of multi-modal (vision-language) LLMs and input the visualized time series plots as a part of the prompt~\cite{wang2024comprehensive,russell2024aad,alnegheimish2024can,liu2024large,liu2025large}. The choice of visual representation over numerical input is motivated by three reasons. First, literature demonstrates that LLMs often struggle with numerical precision due to tokenization artifacts \cite{thawani2021representing,spathis2024first}, where numerical values may be split across tokens or mapped to similar embeddings, introducing potential errors. Second, visual anomaly detection aligns with human cognitive process, as domain experts can more easily identify anomalies through pattern recognition from visualized plots than numerical series. Third, the image-based approach provides greater generalizability across diverse time series metrics without requiring metric-specific pre-processing. 

In addition to the visualized time series data, the business context and domain knowledge are included in the prompt input to the LLMs to ensure that the detection results are aware of the application context. For the output detection results, we specifically require in the prompt that the LLMs should provide with both 1) the binary detection classification (anomalous/normal) and 2) the reason for this detection decision. In this manner, human-level domain knowledge is encoded into the dataset labeled by multimodal LLMs.

To ensure reliable labeling, we implement a two-tier consensus mechanism. We employ multiple multimodal LLMs from different providers to detect anomalies. Each LLM independently analyzes each data point multiple times, and we adopt the majority vote as that model's decision. We only accept a data point's label when all LLMs' majority votes agree, thereby achieving both intra- and inter-model consistency. This multi-model approach with internal voting mitigates potential systematic biases inherent to any single LLM while reducing the impact of stochastic variations in LLM responses\footnote{The prompt input to different multimodal LLMs remains the same and the temperature is set to 0.}.

Although LLMs showcase promising capability in anomaly detection \cite{zhou2024can,dong2024can}, our evaluation reveals that certain data patterns consistently produce disagreement between different models or different trials for an identical model, particularly in boundary cases where the distinction between normal and anomalous behaviors depends on subtle contextual factors. To ensure the quality of the labeled dataset, we implement an iterative refinement process where human experts manually review samples of the labeled data, identify systematic inconsistencies, refine the prompts accordingly to align detection results with business context, and decide those baseline filtering rules to detect anomalies. On the other hand, this does not indicate that human experts can accomplish the labeling task without the help of multimodal LLMs.

Specifically, employing LLMs to automate and standardize the detection procedure helps human experts consolidate the domain expertise: by checking both agreements and disagreements provided by LLMs on large-scale datasets, humans have a more targeted direction to solidify the understandings of anomalies. For example, in initial rounds, LLMs provided inconsistent results when the value of time series moderately rise following long-period zero values. This finding helped humans to set up a deterministic threshold for time series data to be considered as an anomaly. Without LLMs, human experts could have difficulty manually identifying the necessity of specifying such a threshold. In other words, LLMs act as \emph{operators}, who continuously label data points at large scales under instructions, while human experts serve as \emph{managers} to evaluate the results and modify the instructions. This \emph{human-AI collaboration} is essential for generating 1) explicit filtration rules excluding scenarios that LLMs can hardly detect and 2) a high-quality labeled dataset for the following stage.




\subsection{Learning Stage}
\label{sec:learn}
In the learning stage, we first employ the LLM to generate an initial rule prototype (Sec. \ref{sec.generation}). The input to the LLM includes 1) LLM-extracted/summarized reasoning patterns and 2) data-driven statistics, both from the labeled dataset. Then, we implement a systematic iterative improvement pipeline, which improves rules performance through behavioral analysis and targeted modifications, driven by the LLM (Sec. \ref{sec.refinement}). Additionally, the entire learning procedure is consistent with the well-established machine learning pipeline, and therefore ensures generalizability and prevents overfitting (Sec. \ref{sec.complete}). Compared to the first labeling stage involving human-AI collaboration, this learning stage is data-driven and fully automated \emph{without} any human intervention.


\subsubsection{Rule Prototype Generation}
\label{sec.generation}
The learning stage starts from the labeled dataset that contains (i) time series data, (ii) binary anomaly label, and (iii) the reason for the detected label in text form. Both (ii) and (iii) are provided by multimodal LLMs in the previous labeling stage. Our approach first extracts both \textit{qualitative} and \textit{quantitative} information for logic-based symbolic rules generation:
\begin{itemize}
    \item\textbf{Qualitative Reasoning:} We use the LLM to extract the textual reasons from the dataset on why each time series data is detected as an anomaly or not. Extracted textual information includes ``current/recent weeks,'' ``significantly higher than,'' ``breaking established trends,'' etc. Note that we do not include any specific information about the metric nor anomaly descriptions in the prompt during this step, and these qualitative information is inherited from the ``reasons'' in the dataset provided by the multimodal LLMs in the labeling stage.
     \item \textbf{Quantitative Thresholds:} We implement a standard feature engineering procedure to summarize the statistics of normal points and anomalies, including historical statistics, z-scores, trend detection, and zero-rate patterns. This step quantifies the features of normal points/anomalies, where we use standardized Python code instead of LLMs.
\end{itemize}

With both the qualitative and quantitative information provided as the input, a reasoning LLM is invoked to generate the logic-based rules to detect anomalies, with a specified requirement on the output format. Thus we get a logic-based rules prototype for further improvement. This strategy integrates the semantic understanding of LLMs with standard statistical analysis to automatically generate interpretable, executable logic rules without requiring domain expertise or manual feature engineering.


\subsubsection{Iterative Rule Improvement}
\label{sec.refinement}

After obtaining the initial rules prototype, we implement a systematic and automated improvement pipeline to iteratively improve the rules' performance (F1 score for anomaly detection). The improvement process consists of iterative 1) performance evaluation \& behavioral analysis and 2) targeted modification:
\begin{enumerate}
        \item \textbf{Performance Evaluation \& Behavioral Analysis:} In each iteration, the rule is first evaluated by metrics including precision, recall, and F1-score, based on the labeled dataset, with robust handling of edge cases such as rules execution failures. Based on the evaluation, the rules behavioral patterns are categorized into distinct types including \textit{over-conservative}, \textit{over-aggressive}, etc. This behavioral categorization is derived from confusion matrix analysis rather than relying solely on F1-score, providing more granular insights into the rules specific detection deficiencies.

\item 

\textbf{Targeted Modification:} For each identified behavioral pattern, we leverage the LLM's reasoning capabilities to generate targeted rules modifications. The LLM receives 1) the current rules, and 2) behavioral analysis results. Therefore, LLM identifies specific improvement directions corresponding to the detected pattern (e.g., decreasing thresholds for over-conservative rules). The LLM reasons about the logical structure of the rules and generates contextually appropriate rule modifications that address the specific performance deficiencies identified through behavioral analysis.
\end{enumerate}

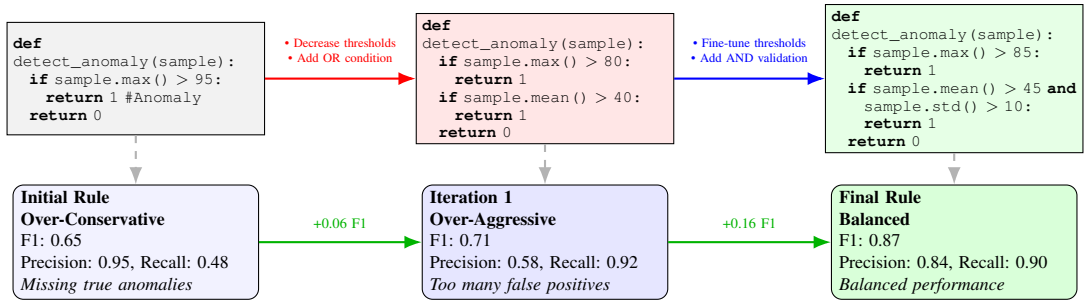
\begin{figure*}[!ht]
\centering
\begin{tikzpicture}[scale=0.68, transform shape,
    rulebox/.style={draw, rounded corners, fill=blue!5, minimum width=4.8cm, minimum height=1.6cm, text width=4.5cm, align=left},
    rulecode/.style={draw, fill=gray!10, font=\ttfamily\small, text width=4.8cm, align=left, minimum height=2.2cm},
    arrow/.style={-{Latex}, thick},
    annot/.style={font=\scriptsize, align=center}
]
\node[rulecode] (rule1) at (0, 4) {
\textbf{def} detect\_anomaly(sample):\\
\quad \textbf{if} sample.max() $>$ 95:\\
\quad\quad \textbf{return} 1 \text{\#Anomaly}\\
\quad \textbf{return} 0
};
\node[rulecode, fill=red!10] (rule2) at (8, 4) {
\textbf{def} detect\_anomaly(sample):\\
\quad \textbf{if} sample.max() $>$ 80:\\
\quad\quad \textbf{return} 1\\
\quad \textbf{if} sample.mean() $>$ 40:\\
\quad\quad \textbf{return} 1\\
\quad \textbf{return} 0
};
\node[rulecode, fill=green!10] (rule3) at (16, 4) {
\textbf{def} detect\_anomaly(sample):\\
\quad \textbf{if} sample.max() $>$ 85:\\
\quad\quad \textbf{return} 1\\
\quad \textbf{if} sample.mean() $>$ 45 \textbf{and}\\
\quad\quad sample.std() $>$ 10:\\
\quad\quad \textbf{return} 1\\
\quad \textbf{return} 0
};
\draw[arrow, red, thick] (rule1.east) -- (rule2.west) 
    node[midway, above=2mm, annot, text=red] {• Decrease thresholds\\• Add OR condition};
    
\draw[arrow, blue, thick] (rule2.east) -- (rule3.west) 
    node[midway, above=2mm, annot, text=blue] {• Fine-tune thresholds\\• Add AND validation};
\node[rulebox] (perf1) at (0, 0.8) {
\textbf{Initial Rule}\\
\textbf{Over-Conservative}\\
F1: 0.65\\
Precision: 0.95, Recall: 0.48\\
\emph{Missing true anomalies}
};
\node[rulebox, fill=blue!10] (perf2) at (8, 0.8) {
\textbf{Iteration 1}\\
\textbf{Over-Aggressive}\\
F1: 0.71\\
Precision: 0.58, Recall: 0.92\\
\emph{Too many false positives}
};
\node[rulebox, fill=green!15] (perf3) at (16, 0.8) {
\textbf{Final Rule}\\
\textbf{Balanced}\\
F1: 0.87\\
Precision: 0.84, Recall: 0.90\\
\emph{Balanced performance}
};
\draw[arrow, thick, green!70!black] (perf1.east) -- (perf2.west) 
    node[midway, above=1.5mm, text=green!70!black, annot, font=\footnotesize] {+0.06 F1};
    
\draw[arrow, thick, green!70!black] (perf2.east) -- (perf3.west) 
    node[midway, above=1.5mm, text=green!70!black, annot, font=\footnotesize] {+0.16 F1};
\draw[arrow, dashed, gray!60] (rule1.south) -- (perf1.north);
\draw[arrow, dashed, gray!60] (rule2.south) -- (perf2.north);
\draw[arrow, dashed, gray!60] (rule3.south) -- (perf3.north);
\end{tikzpicture}
\caption{A simplified illustration of iterative rule improvements: The performance evaluation based on the labeled dataset identifies behavioral patterns of the rule (over-conservative, over-aggressive, etc.) and LLMs apply targeted modifications to refine the rule, achieving the target performance.}
\label{fig:code-evolution}
\end{figure*}

Additionally, the improvement employs a \emph{trajectory-aware} learning procedure. That is, the pipeline maintains a complete history of all improvement attempts (as well as failures), including the rule versions, performance metrics, and modification outcomes. This trajectory information is fed back to the LLM in subsequent iterations, enabling it to learn from previous failures and avoid repeating ineffective modification patterns. The process continues iteratively until the maximum number of iterations is reached. The procedure is formalized and summarized in Algorithm \ref{alg:rule_refinement}, as well as a simplified illustration in Fig. \ref{fig:code-evolution}.

\begin{algorithm}
\caption{Iterative Rule Refinement Pipeline}
\label{alg:rule_refinement}
\begin{algorithmic}[1]
\REQUIRE Initial rule prototype $R_0$, labeled dataset $D$, maximum iterations $N_{max}$
\ENSURE Improved rule $R_{best}$ with performance metrics

\STATE Initialize $R_{current} \leftarrow R_0$, $R_{best} \leftarrow R_0$, trajectory $T \leftarrow \emptyset$
\STATE Evaluate initial performance: $M_0 \leftarrow \text{EvaluateRule}(R_0, D)$
\STATE $M_{current} \leftarrow M_0$, $M_{best} \leftarrow M_0$

\FOR{$i = 1$ to $N_{max}$}
    \STATE \textbf{// Performance Evaluation \& Behavioral Analysis}
    \STATE Compute confusion matrix from $M_{current}$
    \STATE $B \leftarrow \text{AnalyzeBehavior}(M_{current})$ // Categorize: over-conservative, over-aggressive, etc.
    
    \STATE \textbf{// Trajectory-Aware Rule Modification}
    \STATE Update trajectory: $T \leftarrow T \cup \{(R_{current}, M_{current}, B)\}$
    \STATE Generate improvement prompt with $(R_{current}, B, T)$
    \STATE $R_{new} \leftarrow \text{LLM}(\text{prompt})$ // Generate improved rule code
    
    \STATE \textbf{// Evaluation \& Rule Tracking}
    \STATE $M_{new} \leftarrow \text{EvaluateRule}(R_{new}, D)$
    \IF{$M_{new}.\text{F1\_score} > M_{best}.\text{F1\_score}$}
        \STATE $R_{best} \leftarrow R_{new}$, $M_{best} \leftarrow M_{new}$
    \ENDIF
    
    \STATE \textbf{// Update Current Rule for Next Iteration}
    \IF{$M_{new}.\text{F1\_score} > M_{current}.\text{F1\_score}$}
        \STATE $R_{current} \leftarrow R_{new}$, $M_{current} \leftarrow M_{new}$
    \ENDIF
\ENDFOR

\RETURN $R_{best}$, $M_{best}$, trajectory $T$
\end{algorithmic}
\end{algorithm}

This automated improvement approach enables systematic rule optimization through iterative behavioral analysis and targeted modifications, eliminating the need for manual rule tuning while ensuring consistent improvement toward the desired performance target. The trajectory-aware mechanism prevents cyclical improvements and promotes convergent optimization behavior toward the specified performance target. Additionally, the iterative improvement pipeline using LLMs shares some similar spirits with the \emph{automatic heuristic design}, where LLMs automatically propose new algorithms, implement them, get the evaluation feedback, and then revise them based on the feedback \cite{liuevolution,romera2024mathematical,agliettifunbo,ma2024llm,zhang2025socrates}. The proposed algorithms in these methods are largely maintained and decided by a heuristic strategy (e.g,. evolution computation) to complete the entire procedure. In comparison, we cast the LLM-assisted rule learning into a standardized machine learning pipeline.


\subsubsection{Complete Learning Framework for Logic-based Rules}
\label{sec.complete}
During the iterative rule improvement pipeline, the invoked LLM revises the current logic-based rules using the provided ``direction'' (e.g., decreasing the threshold), which is derived from the behavioral analysis (e.g., over-conservative) in the performance evaluation. This ``direction'' shares similar spirit with the gradient descent when minimizing the loss function during the machine learning (ML) procedure \cite{yang2023large}. Inspired by this similarity between our LLM-assisted improvement pipeline and the gradient descent method in classical ML, we develop a systematic procedure to learn a logic-based rule for anomaly detection that follows established learning methodology with training/test splits and early stopping mechanisms. A comprehensive comparison between our proposed method and classical ML learning procedure is included in Table \ref{tab:ml_comparison}.

To be more specific, the complete learning framework for logic-based rules consists of:
\begin{enumerate}
    \item \textbf{Data Split:} The dataset is divided into training and test sets following standard ML practices.


\item \textbf{Two-Phase Learning per Rule:} For each rule, the procedure first generates an initial rule prototype through pattern extraction and feature engineering as in Sec. \ref{sec.generation}. It then randomly divides the training set into learning and validation set, and then conducts multi-epoch learning. In each epoch, the learning procedure implements the iterative improvement pipeline (behavioral analysis → LLM-driven rule modification → performance evaluation) on training data as in Sec. \ref{sec.refinement}, followed by an evaluation on the validation set. The epoch is accepted when the performance gap between the training and validation sets is within a threshold.

\item \textbf{Early Stopping \& Convergence:} For each set of logic-based rules, learning terminates when either the target F1-score is achieved, maximum epochs are reached, or no improvement is observed for a patience period.

\item \textbf{Multiple Rule Generation \& Selection:} The procedure generates multiple rules from the same training dataset to account for the stochastic nature of LLMs. After generating multiple candidate rules, the final rule is selected based on a held-out test dataset, ensuring unbiased model selection and avoiding overfitting.

\end{enumerate}
This learning procedure integrates the interpretability advantages of logic-based rules, supported by LLM, with the well-established learning methodology of classical machine learning, ensuring the generalizability and preventing ovefitting.

\begin{table}[ht]
\centering
\footnotesize
\caption{Comparison of Classical ML \& Our Rule Learning Method}
\label{tab:ml_comparison}
\renewcommand{\arraystretch}{1.2}
\begin{tabular}{@{}lp{0.32\columnwidth}p{0.4\columnwidth}@{}}
\toprule
\textbf{Aspect} & \textbf{Classical ML} & \textbf{Our Method} \\
\midrule
Target & Minimize loss & Maximize F1 score \\
Parameters & Continuous & Discrete rules \\
``Gradient'' & $\nabla L = \partial L/\partial \theta$ & Rule behavior analysis \\
Update & $\theta = \theta - \alpha \nabla L$ & LLM rule modification \\
Rate & Fixed/adaptive & LLM reasoning \\
Memory & Momentum states & Rule trajectory \\
Epoch & Gradient descent & Iterative pipeline \\
Initialization & Random & LLM-generated \\
Convergence & Loss threshold & Target F1 \\
\bottomrule
\end{tabular}
\end{table}

At the end of this part, we explain why we implement LLM-driven logic-based rule learning instead of supervised learning procedures using the labeled data attained in Sec.~\ref{sec:label}, for several key reasons. First, the learned rules are explicit and human-readable, enabling verification and optional revision by human experts. In contrast, machine learning models are generally ``black boxes'' and hard to interpret in production. Second, although LLMs showcase promising anomaly detection capabilities, their detection results are not fully reliable and sometimes suffer from inconsistencies or hallucinations. Logic-based rules learned from the labeled dataset ``distill'' the predominant patterns, while supervised learning methods may overfit to noise in the training data. Lastly, time series data in supply chain management exhibits strong non-stationarity: data during holiday weeks or promotions differs significantly from that in normal weeks. This non-stationarity leads to distribution shifts, under which supervised learning methods degrade while our learned rules, with deterministic and consistent logic, exhibit strong robustness. We include numerical results to support the advantages of our learned rules over supervised learning methods in Sec.~\ref{sec:exp}.

\subsection{Augmentation Stage}
\label{sec:revision}

In this stage, we implement a semantic categorization to enhance the interpretability of the rules through two steps: \textbf{1. Rule Analysis \& Category Taxonomy:} Our method analyzes the learned rules to extract the detection principles using LLMs, identifying strategies such as z-score thresholds, ratio comparisons, and trend detection. Based on the rule analysis, the LLM generates a taxonomy of anomalous categories accounting for business contexts, e.g., ``Critical Stock-Out Crisis,'' ``Moderate Stock Pressure,'' ``Escalating Stock-Out Emergency.'' \textbf{2. Rule Enhancement \& Category Assignment:} The leaned rules are enriched to classify anomalies into the identified categories without modifying its detection logic, supported by the LLM. The enhanced rules return both the detection results and the category classifications, enabling more informed analysis; see an illustrative example mapping from rules to categories in Table  \ref{tab:rule_categories}.

This categorization enhancement provides critical context for human analysts by explaining not just \textit{that} an anomaly occurred, but \textit{what type} of anomaly was detected. Importantly, this semantic layer is added without sacrificing detection performance, as the underlying detection logic remains unchanged. The output categorization enables more targeted remediation strategies and improved root cause analysis by distinguishing between different categories of anomalies.

\begin{table}[ht]
\centering
\small
\caption{Rule Examples and Their Categories}
\label{tab:rule_categories}
\renewcommand{\arraystretch}{1.3}
\begin{tabular}{@{}p{0.32\columnwidth}|p{0.61\columnwidth}@{}}
\toprule
\textbf{Category} & \textbf{Rules of Anomalies} \\
\midrule
\multirow{3}{=}{\textbf{Critical Stock-Out Crisis}} 
& \texttt{current\_value >= 80.0} \\
\arrayrulecolor{lightgray}\cline{2-2}\arrayrulecolor{black}
& \texttt{ratio of current\_value and values[-2]> 10 \&\& current\_value >= 50} \\
\midrule
\multirow{2}{=}{\textbf{Moderate Stock Pressure}} 
& \texttt{current\_value >= 35 \&\& z\_score >= 5.0} \\
\arrayrulecolor{lightgray}\cline{2-2}\arrayrulecolor{black}
& \texttt{recent\_mean < 3.0 \&\& current\_value >= 12} \\
\bottomrule
\end{tabular}
\end{table}

\begin{table*}
\centering
\caption{Performance comparison of anomaly detection methods on 10K ASINs (892 anomalies)}
\label{tab:performance_comparison}
\begin{tabular}{lccccc}
\toprule
\textbf{Method} & \textbf{Recall (\%)} & \textbf{Precision (\%)} & \textbf{F1 Score (\%)} & \makecell{\textbf{Execution} \\ \textbf{Time (s)}} & \makecell{\textbf{Business} \\ \textbf{Interpretability}} \\
\midrule
iForest-Based Method & 83.95 & 80.03 & 81.94 & 1,250.33 & Medium \\
\midrule
LSTM-VAE Hybrid & 88.00 & 66.98 & 76.04 & 348.50 & Low \\
Anomaly Transformer & 90.02 & 72.02 & 79.93 & 425.80 & Low \\
\midrule
Claude Sonnet 4 & 94.96 & 97.02 & 95.98 & 18,234.17 & High \\
Amazon Nova Pro & 93.05 & 94.97 & 94.00 & 5,267.28 & High \\
Meta Llama 3.2 & 92.04 & 93.94 & 92.98 & 4,838.37 & High \\
\midrule
\textbf{Logic-Based Rules} & \textbf{91.03} & \textbf{93.01} & \textbf{92.01} & \textbf{4.27} & \textbf{High} \\
\bottomrule
\end{tabular}
\label{table:comparison}
\end{table*}

\section{Experiment Results}
\label{sec:exp}
We present the numerical results based on weekly data that tracks stock health for Amazon's individual products (ASINs). Due to business confidentiality, we do not include detailed descriptions of this metric here. Typically, a relatively high value of this metric is flagged as an anomaly, though recent trends and historical value ranges must also be considered. The task is to determine whether the current week (the last timestamp) indicates an anomaly for each ASIN. In production, this anomaly detection system scales to process hundreds of millions of individual products weekly using real data. The evaluation is based on a dataset containing 10K ASINs with 53 weeks of data ($\approx$1 year). The dataset is labeled by multimodal (vision-language) LLMs and reviewed by human experts as described in Sec. \ref{sec:label}. To avoid data leakage, the evaluation dataset has no overlap with the dataset used for learning the logic-based rules.

The compared baseline methods include:
\begin{itemize}
    \item Amazon's in-production anomaly detection at aggregated category levels, which is a modified Isolation Forest (iForest) \cite{4781136} with business context manually encoded.
    \item Unsupervised deep learning models including VAE-LSTM Hybrid \cite{lin2020anomaly}, and Anomaly Transformer \cite{xuanomaly}.
    \item Direct LLMs using the same input as in the labeling stage (Sec. \ref{sec:label}), i.e., we reproduce the results. The employed multimodal LLMs include 1) Anthropic Claude Sonnet 4, 2) Amazon Nova Pro, and 3) Meta Llama 3.2.
\end{itemize}

\begin{figure}
    \centering
    \includegraphics[width=\linewidth]{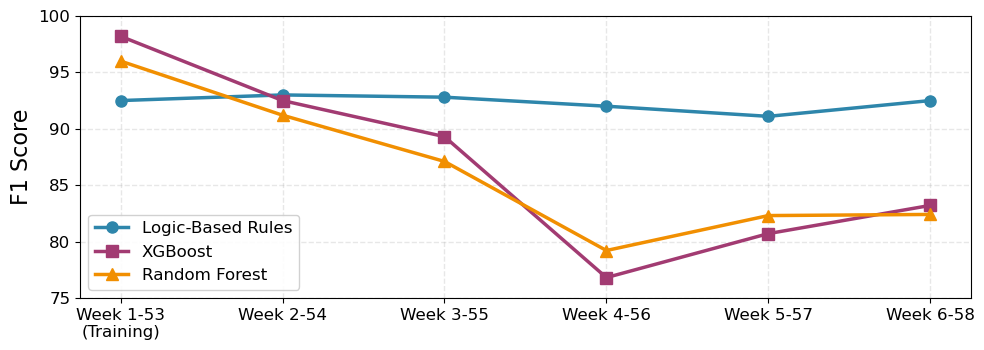}
    \caption{The comparison between the learned rules and supervised learning models trained on the label dataset. Week 56 includes a holiday that causes demand spikes across products, degrading the performance of supervised learning models because of the high anomaly rate.}
    \label{fig:enter-label}
\end{figure}

The results are included in Table \ref{table:comparison}, where the execution time represents the computation cost for detecting anomalies across the entire dataset (10K ASINs). The implementation is hosted on an AWS SageMaker ml.r7i.48xlarge instance type. The experimental results provide the following insights:
\begin{enumerate}
    \item Deep learning methods achieve unsatisfactory results due to low precision, demonstrating over-sensitivity. Specifically, the business context reveals that some trend-breaking patterns (e.g., sharp decreases) are not actual anomalies, yet these unsupervised methods lack awareness of this context, resulting in false positives.
    \item 
Amazon's in-production anomaly detection method (based on iForest) is specifically designed for aggregated category levels, achieving balanced performance with both reasonable precision and recall, since business context has been \emph{manually} encoded into the algorithm. However, its performance still degrades at the individual ASIN level since time series data exhibit significantly higher uncertainty and volatility, thus providing minimal detectable patterns for iForest to leverage.
\item Multimodal LLMs are required to \emph{reproduce} their detection results. Although the detection accuracy of using LLMs directly outperforms other methods, the results demonstrate that the detection outputs are not fully reproducible. This non-deterministic nature creates risks of missed alarms or wasted efforts in production. Additionally, the computational latency and costs associated with LLMs are critical concerns, since Amazon requires weekly anomaly detection across millions of ASINs.
\end{enumerate}
Overall, our learned logic-based rules outperform other baseline methods and have been deployed in production. Additionally, we include an ablation study that compares 1) our logic-based rule learning approach (Sec.\ref{sec:learn}) and 2) a supervised learning procedure using the labeled dataset. Specifically, upon completion of the labeling stage (Sec.~\ref{sec:label}), each data point in the ASIN time series is assigned a binary label (anomalous or normal). Thus, the original unsupervised time series anomaly detection problem is reformulated as a binary classification problem that supervised learning methods can address. We evaluated different models; XGBoost achieved the highest accuracy, followed closely by Random Forest. We evaluate approaches using a rolling 53-week window: the initial dataset for training spans Week 1 to Week 53, and the first evaluation window spans Week 2 to Week 54, with subsequent windows shifted by one week for a total of 5 evaluations. Comparison results among our logic-based rules, XGBoost, and Random Forest are shown in Fig.~\ref{fig:enter-label}. The results demonstrate the robustness advantage of our rules against changing markets (distribution shifts) compared to supervised learning methods, which are sensitive to the underlying anomaly distribution.

\section{Conclusion}

We propose a novel framework that leverages large language models (LLMs) to systematically transform implicit human expertise into explicit logic-based symbolic rules for time series anomaly detection in large-scale supply chain management. Our three-stage methodology comprises LLM-based data labeling, LLM-driven rule generation with iterative improvements, and LLM-assisted semantic categorization augmentation. Experimental results validate the superiority of our approach over baseline methods. Beyond its immediate application to supply chain metrics, our framework represents a generalizable strategy for domains requiring the scaling of human expertise to massive operations. By using LLMs as a bridge to extract and formalize domain knowledge into deterministic rules, this approach enables efficient and interpretable detection systems applicable to various scenarios.

\newpage
\bibliographystyle{IEEEtran}
\bibliography{references}

@inproceedings{zhang2025socrates,
  title={SOCRATES: Simulation Optimization with Correlated Replicas and Adaptive Trajectory Evaluations},
  author={Zhang, Haoting and Chen, Haoxian and Zhan, Donglin and Zhao, Hanyang and Lam, Henry and Tang, Wenpin and Yao, David and Zheng, Zeyu},
  year={2025},
  booktitle={NeurIPS 2025 Workshop MLxOR: Mathematical Foundations and Operational Integration of Machine Learning for Uncertainty-Aware Decision-Making}
}

@inproceedings{zhan2024meta,
  title={Meta-adaptive stock movement prediction with two-stage representation learning},
  author={Zhan, Donglin and Dai, Yusheng and Dong, Yiwei and He, Jinghai and Wang, Zhenyi and Anderson, James},
  booktitle={Proceedings of the 2024 SIAM International Conference on Data Mining (SDM)},
  pages={508--516},
  year={2024},
  organization={SIAM}
}

@article{zhang2024does,
  title={Does Attention in Transformers Help Wildfire Prediction?},
  author={Zhang, Yunkai and Zhan, Donglin and Zhang, Haoting and Shen, Zuo-Jun Max and Zheng, Zeyu and Zhu, Qing},
  journal={Available at SSRN 4929315},
  year={2024}
}

@inproceedings{zhang2024daily,
  title={Daily Physical Activity Monitoring: Adaptive Learning from Multi-source Motion Sensor Data},
  author={Zhang, Haoting and Zhan, Donglin and Lin, Yunduan and He, Jinghai and Zhu, Qing and Shen, Zuo-Jun and Zheng, Zeyu},
  booktitle={Conference on Health, Inference, and Learning},
  pages={39--54},
  year={2024},
  organization={PMLR}
}

@inproceedings{lin2020anomaly,
  title={Anomaly detection for time series using vae-lstm hybrid model},
  author={Lin, Shuyu and Clark, Ronald and Birke, Robert and Sch{\"o}nborn, Sandro and Trigoni, Niki and Roberts, Stephen},
  booktitle={ICASSP 2020-2020 IEEE International Conference on Acoustics, Speech and Signal Processing (ICASSP)},
  pages={4322--4326},
  year={2020},
  organization={Ieee}
}

@inproceedings{sun2024unraveling,
  title={Unraveling the ‘anomaly’in time series anomaly detection: A self-supervised tri-domain solution},
  author={Sun, Yuting and Pang, Guansong and Ye, Guanhua and Chen, Tong and Hu, Xia and Yin, Hongzhi},
  booktitle={2024 IEEE 40th International Conference on Data Engineering (ICDE)},
  pages={981--994},
  year={2024},
  organization={IEEE}
}

@inproceedings{fang2024temporal,
  title={Temporal-frequency masked autoencoders for time series anomaly detection},
  author={Fang, Yuchen and Xie, Jiandong and Zhao, Yan and Chen, Lu and Gao, Yunjun and Zheng, Kai},
  booktitle={2024 IEEE 40th International Conference on Data Engineering (ICDE)},
  pages={1228--1241},
  year={2024},
  organization={IEEE}
}

@inproceedings{xuanomaly,
  title={Anomaly Transformer: Time Series Anomaly Detection with Association Discrepancy},
  author={Xu, Jiehui and Wu, Haixu and Wang, Jianmin and Long, Mingsheng},
  booktitle={International Conference on Learning Representations},
year = {2022}
}

@INPROCEEDINGS{4781136,
  author={Liu, Fei Tony and Ting, Kai Ming and Zhou, Zhi-Hua},
  booktitle={2008 Eighth IEEE International Conference on Data Mining}, 
  title={Isolation Forest}, 
  year={2008},
  volume={},
  number={},
  pages={413-422},
  doi={10.1109/ICDM.2008.17}}

@article{thawani2021representing,
  title={Representing numbers in NLP: a survey and a vision},
  author={Thawani, Avijit and Pujara, Jay and Szekely, Pedro A and Ilievski, Filip},
  journal={arXiv preprint arXiv:2103.13136},
  year={2021}
}

@article{spathis2024first,
  title={The first step is the hardest: Pitfalls of representing and tokenizing temporal data for large language models},
  author={Spathis, Dimitris and Kawsar, Fahim},
  journal={Journal of the American Medical Informatics Association},
  volume={31},
  number={9},
  pages={2151--2158},
  year={2024},
  publisher={Oxford Academic}
}

@article{wang2024comprehensive,
  title={A comprehensive review of multimodal large language models: Performance and challenges across different tasks},
  author={Wang, Jiaqi and Jiang, Hanqi and Liu, Yiheng and Ma, Chong and Zhang, Xu and Pan, Yi and Liu, Mengyuan and Gu, Peiran and Xia, Sichen and Li, Wenjun and others},
  journal={arXiv preprint arXiv:2408.01319},
  year={2024}
}

@article{zamanzadeh2024deep,
  title={Deep learning for time series anomaly detection: A survey},
  author={Zamanzadeh Darban, Zahra and Webb, Geoffrey I and Pan, Shirui and Aggarwal, Charu and Salehi, Mahsa},
  journal={ACM Computing Surveys},
  volume={57},
  number={1},
  pages={1--42},
  year={2024},
  publisher={ACM New York, NY}
}

@article{paparrizos2022volume,
  title={Volume under the surface: a new accuracy evaluation measure for time-series anomaly detection},
  author={Paparrizos, John and Boniol, Paul and Palpanas, Themis and Tsay, Ruey S and Elmore, Aaron and Franklin, Michael J},
  journal={Proceedings of the VLDB Endowment},
  volume={15},
  number={11},
  pages={2774--2787},
  year={2022},
  publisher={VLDB Endowment}
}

@inproceedings{liang2025kdselector,
  title={KDSelector: A Knowledge-Enhanced and Data-Efficient Model Selector Learning Framework for Time Series Anomaly Detection},
  author={Liang, Zhiyu and Cai, Dongrui and Zhang, Chenyuan and Liang, Zheng and Liang, Chen and Zheng, Bo and Qiu, Shi and Wang, Jin and Wang, Hongzhi},
  booktitle={Companion of the 2025 International Conference on Management of Data},
  pages={175--178},
  year={2025}
}

@article{schneider2021distributed,
  title={Distributed detection of sequential anomalies in univariate time series},
  author={Schneider, Johannes and Wenig, Phillip and Papenbrock, Thorsten},
  journal={The VLDB Journal},
  volume={30},
  number={4},
  pages={579--602},
  year={2021},
  publisher={Springer}
}

@article{liu2025easyad,
  title={EasyAD: A Demonstration of Automated Solutions for Time-Series Anomaly Detection},
  author={Liu, Qinghua and Lee, Seunghak and Paparrizos, John},
  journal={Proceedings of the VLDB Endowment},
  volume={18},
  number={12},
  pages={5431--5434},
  year={2025},
  publisher={VLDB Endowment}
}

@article{ahmed2021anomaly,
  title={Anomaly detection, localization and classification using drifting synchrophasor data streams},
  author={Ahmed, Arman and Sajan, K Sadanandan and Srivastava, Anurag and Wu, Yinghui},
  journal={IEEE Transactions on Smart Grid},
  volume={12},
  number={4},
  pages={3570--3580},
  year={2021},
  publisher={IEEE}
}

@article{chen2023imdiffusion,
  title={ImDiffusion: Imputed Diffusion Models for Multivariate Time Series Anomaly Detection},
  author={Chen, Yuhang and Zhang, Chaoyun and Ma, Minghua and Liu, Yudong and Ding, Ruomeng and Li, Bowen and He, Shilin and Rajmohan, Saravan and Lin, Qingwei and Zhang, Dongmei},
  journal={Proceedings of the VLDB Endowment},
  volume={17},
  number={3},
  pages={359--372},
  year={2023},
  publisher={VLDB Endowment}
}

@article{romera2024mathematical,
  title={Mathematical discoveries from program search with large language models},
  author={Romera-Paredes, Bernardino and Barekatain, Mohammadamin and Novikov, Alexander and Balog, Matej and Kumar, M Pawan and Dupont, Emilien and Ruiz, Francisco JR and Ellenberg, Jordan S and Wang, Pengming and Fawzi, Omar and others},
  journal={Nature},
  volume={625},
  number={7995},
  pages={468--475},
  year={2024},
  publisher={Nature Publishing Group UK London}
}

@inproceedings{agliettifunbo,
  title={FunBO: Discovering Acquisition Functions for Bayesian Optimization with FunSearch},
  author={Aglietti, Virginia and Ktena, Ira and Schrouff, Jessica and Sgouritsa, Eleni and Ruiz, Francisco and Malek, Alan and Bellot, Alexis and Chiappa, Silvia},
  booktitle={Forty-second International Conference on Machine Learning},
  year={2024}
}

@inproceedings{ma2024llm,
  title={LLM and Simulation as Bilevel Optimizers: A New Paradigm to Advance Physical Scientific Discovery},
  author={Ma, Pingchuan and Wang, Tsun-Hsuan and Guo, Minghao and Sun, Zhiqing and Tenenbaum, Joshua B and Rus, Daniela and Gan, Chuang and Matusik, Wojciech},
  booktitle={International Conference on Machine Learning},
  pages={33940--33962},
  year={2024},
  organization={PMLR}
}

@inproceedings{liuevolution,
  title={Evolution of Heuristics: Towards Efficient Automatic Algorithm Design Using Large Language Model},
  author={Liu, Fei and Xialiang, Tong and Yuan, Mingxuan and Lin, Xi and Luo, Fu and Wang, Zhenkun and Lu, Zhichao and Zhang, Qingfu},
  booktitle={Forty-first International Conference on Machine Learning},
  year={2024}
}

@article{zhou2025mtsbench,
  title={mTSBench: Benchmarking Multivariate Time Series Anomaly Detection and Model Selection at Scale},
  author={Zhou, Xiaona and Brif, Constantin and Lourentzou, Ismini},
  journal={arXiv preprint arXiv:2506.21550},
  year={2025}
}

@article{yao2025moon,
  title={Moon: A Modality Conversion-based Efficient Multivariate Time Series Anomaly Detection},
  author={Yao, Yuanyuan and Shi, Yuhan and Chen, Lu and Fang, Ziquan and Gao, Yunjun and Li, Yushuai and Li, Tianyi and others},
  journal={arXiv preprint arXiv:2510.01970},
  year={2025}
}

@article{paparrizos2022tsb,
  title={TSB-UAD: an end-to-end benchmark suite for univariate time-series anomaly detection},
  author={Paparrizos, John and Kang, Yuhao and Boniol, Paul and Tsay, Ruey S and Palpanas, Themis and Franklin, Michael J},
  journal={Proceedings of the VLDB Endowment},
  volume={15},
  number={8},
  pages={1697--1711},
  year={2022},
  publisher={VLDB Endowment}
}

@article{srivastava2020forecasting,
  title={Forecasting in multivariate irregularly sampled time series with missing values},
  author={Srivastava, Shivam and Sen, Prithviraj and Reinwald, Berthold},
  journal={arXiv preprint arXiv:2004.03398},
  year={2020}
}

@article{liu2025tsb,
  title={Tsb-autoad: Towards automated solutions for time-series anomaly detection},
  author={Liu, Qinghua and Lee, Seunghak and Paparrizos, John},
  journal={Proceedings of the VLDB Endowment},
  volume={18},
  number={11},
  pages={4364--4379},
  year={2025},
  publisher={VLDB Endowment}
}

@inproceedings{zhou2025crosslinear,
  title={CrossLinear: Plug-and-Play Cross-Correlation Embedding for Time Series Forecasting with Exogenous Variables},
  author={Zhou, Pengfei and Liu, Yunlong and Liang, Junli and Song, Qi and Li, Xiangyang},
  booktitle={Proceedings of the 31st ACM SIGKDD Conference on Knowledge Discovery and Data Mining V. 2},
  pages={4120--4131},
  year={2025}
}

@inproceedings{jhin2023precursor,
  title={Precursor-of-anomaly detection for irregular time series},
  author={Jhin, Sheo Yon and Lee, Jaehoon and Park, Noseong},
  booktitle={Proceedings of the 29th ACM SIGKDD Conference on Knowledge Discovery and Data Mining},
  pages={917--929},
  year={2023}
}

@inproceedings{yang2023dcdetector,
  title={Dcdetector: Dual attention contrastive representation learning for time series anomaly detection},
  author={Yang, Yiyuan and Zhang, Chaoli and Zhou, Tian and Wen, Qingsong and Sun, Liang},
  booktitle={Proceedings of the 29th ACM SIGKDD conference on knowledge discovery and data mining},
  pages={3033--3045},
  year={2023}
}

@article{han2022deeptea,
  title={DeepTEA: Effective and efficient online time-dependent trajectory outlier detection},
  author={Han, Xiaolin and Cheng, Reynold and Ma, Chenhao and Grubenmann, Tobias},
  journal={Proceedings of the VLDB Endowment},
  volume={15},
  number={7},
  pages={1493--1505},
  year={2022},
  publisher={VLDB Endowment}
}

@inproceedings{audibert2020usad,
  title={Usad: Unsupervised anomaly detection on multivariate time series},
  author={Audibert, Julien and Michiardi, Pietro and Guyard, Fr{\'e}d{\'e}ric and Marti, S{\'e}bastien and Zuluaga, Maria A},
  booktitle={Proceedings of the 26th ACM SIGKDD international conference on knowledge discovery \& data mining},
  pages={3395--3404},
  year={2020}
}

@article{munir2018deepant,
  title={DeepAnT: A deep learning approach for unsupervised anomaly detection in time series},
  author={Munir, Mohsin and Siddiqui, Shoaib Ahmed and Dengel, Andreas and Ahmed, Sheraz},
  journal={Ieee Access},
  volume={7},
  pages={1991--2005},
  year={2018},
  publisher={IEEE}
}

@article{zhao2022comparative,
  title={A comparative study on unsupervised anomaly detection for time series: Experiments and analysis},
  author={Zhao, Yan and Deng, Liwei and Chen, Xuanhao and Guo, Chenjuan and Yang, Bin and Kieu, Tung and Huang, Feiteng and Pedersen, Torben Bach and Zheng, Kai and Jensen, Christian S},
  journal={arXiv preprint arXiv:2209.04635},
  year={2022}
}

@inproceedings{obata2025robust,
  title={Robust and Explainable Detector of Time Series Anomaly via Augmenting Multiclass Pseudo-Anomalies},
  author={Obata, Kohei and Matsubara, Yasuko and Sakurai, Yasushi},
  booktitle={Proceedings of the 31st ACM SIGKDD Conference on Knowledge Discovery and Data Mining V. 2},
  pages={2198--2209},
  year={2025}
}

@inproceedings{liu2025large,
  title={Large language models can deliver accurate and interpretable time series anomaly detection},
  author={Liu, Jun and Zhang, Chaoyun and Qian, Jiaxu and Ma, Minghua and Qin, Si and Bansal, Chetan and Lin, Qingwei and Rajmohan, Saravan and Zhang, Dongmei},
  booktitle={Proceedings of the 31st ACM SIGKDD Conference on Knowledge Discovery and Data Mining V. 2},
  pages={4623--4634},
  year={2025}
}

@article{gu2025argos,
  title={Argos: Agentic time-series anomaly detection with autonomous rule generation via large language models},
  author={Gu, Yile and Xiong, Yifan and Mace, Jonathan and Jiang, Yuting and Hu, Yigong and Kasikci, Baris and Cheng, Peng},
  journal={arXiv preprint arXiv:2501.14170},
  year={2025}
}

@article{dong2024can,
  title={Can LLMs Serve As Time Series Anomaly Detectors?},
  author={Dong, Manqing and Huang, Hao and Cao, Longbing},
  journal={arXiv preprint arXiv:2408.03475},
  year={2024}
}

@article{zhou2024can,
  title={Can LLMs understand time series anomalies?},
  author={Zhou, Zihao and Yu, Rose},
  journal={arXiv preprint arXiv:2410.05440},
  year={2024}
}

@article{tsai2025anollm,
  title={AnoLLM: Large language models for tabular anomaly detection},
  author={Tsai, Che-Ping and Teng, Ganyu and Wallis, Phil and Ding, Wei},
  year={2025}
}

@inproceedings{chandrayanlead,
  title={LEAD-Framework for efficient time-series anomaly detection on large scale data using LLMs},
  author={Chandrayan, Akash and Amir, ZIDI and Reimherr, Matthew and Mjirda, Anis and Pradhan, Abhinav},
  booktitle={1st ICML Workshop on Foundation Models for Structured Data}
}

@article{jin2023time,
  title={Time-llm: Time series forecasting by reprogramming large language models},
  author={Jin, Ming and Wang, Shiyu and Ma, Lintao and Chu, Zhixuan and Zhang, James Y and Shi, Xiaoming and Chen, Pin-Yu and Liang, Yuxuan and Li, Yuan-Fang and Pan, Shirui and others},
  journal={arXiv preprint arXiv:2310.01728},
  year={2023}
}

@article{gruver2023large,
  title={Large language models are zero-shot time series forecasters},
  author={Gruver, Nate and Finzi, Marc and Qiu, Shikai and Wilson, Andrew G},
  journal={Advances in Neural Information Processing Systems},
  volume={36},
  pages={19622--19635},
  year={2023}
}

@inproceedings{liu2024large,
  title={Large language model guided knowledge distillation for time series anomaly detection},
  author={Liu, Chen and He, Shibo and Zhou, Qihang and Li, Shizhong and Meng, Wenchao},
  booktitle={Proceedings of the Thirty-Third International Joint Conference on Artificial Intelligence},
  pages={2162--2170},
  year={2024}
}

@inproceedings{yang2023large,
  title={Large language models as optimizers},
  author={Yang, Chengrun and Wang, Xuezhi and Lu, Yifeng and Liu, Hanxiao and Le, Quoc V and Zhou, Denny and Chen, Xinyun},
  booktitle={The Twelfth International Conference on Learning Representations},
  year={2023}
}

@article{tedjopurnomo2020survey,
  title={A survey on modern deep neural network for traffic prediction: Trends, methods and challenges},
  author={Tedjopurnomo, David Alexander and Bao, Zhifeng and Zheng, Baihua and Choudhury, Farhana Murtaza and Qin, Alex Kai},
  journal={IEEE Transactions on Knowledge and Data Engineering},
  volume={34},
  number={4},
  pages={1544--1561},
  year={2020},
  publisher={IEEE}
}

@inproceedings{russell2024aad,
  title={Aad-llm: Adaptive anomaly detection using large language models},
  author={Russell-Gilbert, Alicia and Sommers, Alexander and Thompson, Andrew and Cummins, Logan and Mittal, Sudip and Rahimi, Shahram and Seale, Maria and Jaboure, Joseph and Arnold, Thomas and Church, Joshua},
  booktitle={2024 IEEE International Conference on Big Data (BigData)},
  pages={4194--4203},
  year={2024},
  organization={IEEE}
}

@inproceedings{alnegheimish2024can,
  title={Can large language models be anomaly detectors for time series?},
  author={Alnegheimish, Sarah and Nguyen, Linh and Berti-Equille, Laure and Veeramachaneni, Kalyan},
  booktitle={2024 IEEE 11th International Conference on Data Science and Advanced Analytics (DSAA)},
  pages={1--10},
  year={2024},
  organization={IEEE}
}

\end{document}